\pdfoutput=1

\documentclass[11pt]{article}

\usepackage[review]{EMNLP2022}

\usepackage{times}
\usepackage{latexsym}

\usepackage[T1]{fontenc}

\usepackage[utf8]{inputenc}

\usepackage{microtype}

\usepackage{inconsolata}
\usepackage{graphicx}
\usepackage{amsfonts,amssymb}
\usepackage{bm}
\usepackage{booktabs}
\usepackage{array}

\title{Wish I Can Feel What You Feel: A Neural Approach for Empathetic Response Generation}

\author{Yangbin Chen \and Chunfeng Liang\\
        Suzhou Fubian Medical Technology Co., Ltd., China \\ 
        {\fontfamily{qcr}\selectfont \{dongyiwu92, cfliang666\}}{\fontfamily{qcr}\selectfont @gmail.com}
                                        }

\begin{document}
\maketitle
\begin{abstract}
Expressing empathy is important in everyday conversations, and exploring how empathy arises is crucial in automatic response generation. 
Most previous approaches consider only a single factor that affects empathy. 
However, in practice, empathy generation and expression is a very complex and dynamic psychological process. 
A listener needs to find out events which cause a speaker's emotions (emotion cause extraction), project the events into some experience (knowledge extension), and express empathy in the most appropriate way (communication mechanism).
To this end, we propose a novel approach, which integrates the three components - emotion cause, knowledge graph, and communication mechanism for empathetic response generation.
Experimental results on the benchmark dataset demonstrate the effectiveness of our method and show that incorporating the key components generates more informative and empathetic responses.  
\end{abstract}

\section{Introduction}
According to \citet{Hoffman2000EmpathyAM}, empathy is an affective response more appropriate to another's situation than one's own, which is the spark of human concern for others and the glue that makes social life possible.
It is a complex human trait and dynamic psychological process related to emotion and cognition, where emotional empathy refers to vicarious sharing of emotion and cognitive empathy refers to mental perspective taking 
\citep{smith2006}. 
Since 1990s, the study of empathy has been widely applied to mental health support \cite{bohart1997,fitzpatrick2017}, quality of care improvement \citep{mercer2002}, and intelligent virtual assistants \citep{shin2019}.

\begin{figure}[htbp] 
 \center{\includegraphics[width=8cm]  {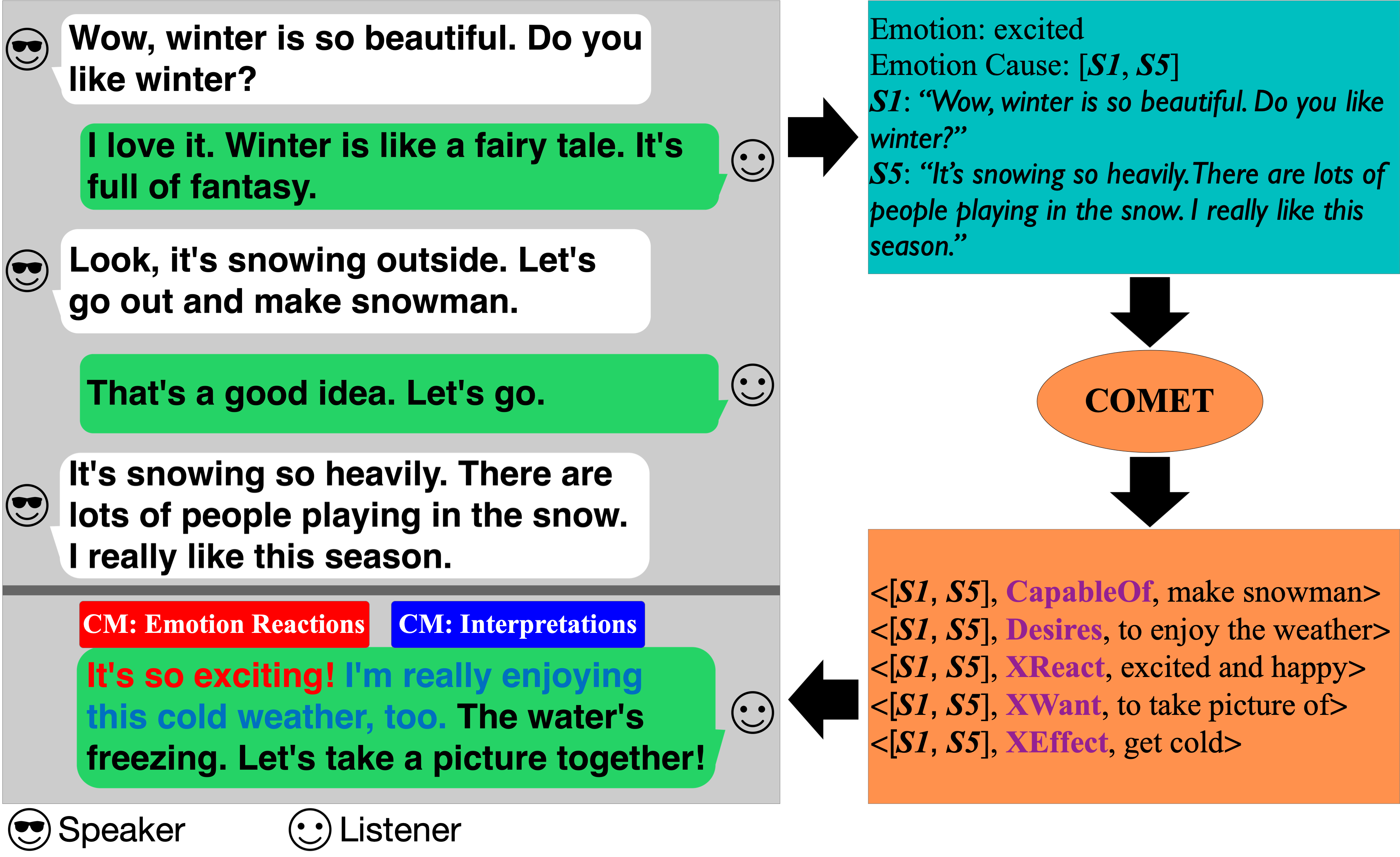}} 
 \caption{An example of empathetic response from EMPATHETICDIALOGUES dataset. In the teal box are emotion and causes detected from the dialogue context. In the orange box is extended knowledge via COMET. The colored texts in the final reply show two types of communication mechanisms.} 
 \label{sec:introduction}
 \end{figure}
 
Expressing empathy becomes more important in today's dialogue systems. 
However, there are challenges in developing an empathetic model, such as preparing a proper training corpus, learning to get a comprehensive understanding of the dialogue context, and designing an appropriate empathy expression strategy.


Recently, there has been some work to address these issues.  
A standard benchmark containing large-scale empathetic conversations was proposed, laying the cornerstone of empathetic dialogue research \cite{rashkin2018}.
Some researchers try to gain a deeper understanding of contextual information.
For example, \citet{gao2021} applied an emotion cause extractor to conversations and used the extracted causes to guide the response generation process.
\citet{li2022} incorporated external commonsense information to enrich the context.
During the language generation process, some researchers focus on controlling emotions of generated responses using emotional blending to imitate the speakers' emotions \cite{majumder2020,lin2019}.

All the above work considers only a single aspect that affects empathy.
However, in practice, empathy generation and expression is a very complex and dynamic process.
According to research work in the field of psychological science, we believe that three different but related factors matter in empathy: emotion (the automatic proclivity to share emotions with others), cognition (the intersubjectivity to interpret others' intentions and feelings while keeping separate self and other perspectives), and behavioral outcome (the actions to express empathy) \cite{decety2008emotion,heyes2018empathy}.
Consequently, we divide the entire empathy process into five functional modules: emotion perception, cause extraction, experience projection, dialogue reaction, and verbal expression.
Specifically, emotion perception aims to sense emotions from others.
Cause extraction is to determine detailed events corresponding to the emotions.
Experience projection enriches the contextual information through knowledge extension from the emotion causes.
Dialogue reaction decides the response strategies by learning from the contexts.
Verbal expression is the final step in a dialogue system to generate responses in terms of languages.

Towards this end, we propose a novel approach \textbf{IMAGINE}, a.k.a. \textbf{I}ntegrating e\textbf{M}otion c\textbf{A}uses, knowled\textbf{G}e, and commun\textbf{I}catio\textbf{N} m\textbf{E}chanisms for empathetic dialogue generation. 
Using these components improves cognitive understanding of contexts and enhances empathy expression in the generated responses.
Our framework involves three stages -- emotion cause extraction, knowledge-enriched communication, and response generation. 
We evaluate our approach on the EMPATHETICDIALOGUES dataset.
Extensive experimental results demonstrate the effectiveness of \textbf{IMAGINE} in automatic and human evaluations, showing that our approach generates more informative and empathetic responses (An example is shown in Figure ${\color{red}{\fbox{\ref{sec:introduction}}}}$).

Our contributions can be summarized as follows:

\indent \textbf{1)} We propose a new approach \textbf{IMAGINE} which integrates emotion causes, knowledge, and communication mechanisms into a dialogue system, demonstrating that they are significant factors in the generation and expression of empathy.

\indent \textbf{2)} We divide relationships within a knowledge graph into several categories, including Affect, Behaviour, Physical, and Events. 
Meanwhile, we design a three-stage process of emotion cause extraction, knowledge-enriched communication, and response generation based on the dialogue history.

\indent \textbf{3)} Experimental results show that our proposed approach significantly outperforms other comparison methods, with more informative and empathetic responses.

\section{Related Work}


\subsection{Empathetic dialogue generation}
Empathetic response generation is a sub-task of emotion-aware response generation.
\citet{rashkin2018} first proposed a standard benchmark containing large-scale empathetic conversations. 
Some researchers focus on understanding the dialogue context.
\citet{li2021} and \citet{gao2021} identified the emotion causes of the conversation to understand the context related to emotions better.
\citet{sabour2021} and \citet{li2022} leveraged external knowledge, including commonsense knowledge and emotional lexical knowledge, to explicitly understand and express emotions.
Some researchers focus on the language generation process, for example, controlling emotions of generated responses through mixture model \cite{lin2019}, adversarial framework \cite{li2019}, and mimicking the emotions of the speaker \cite{majumder2020}.
\citet{sharma2020} and \citet{zheng2021} explore the expressive factors that elicit empathy.
Moreover, as big models are popular today, \citet{lin2020} adapted GPT2 \citep{radford2019} to produce empathetic responses via transfer learning, active learning, and negative training.

\begin{figure*}[htbp] 
\center{\includegraphics[width=16cm]  {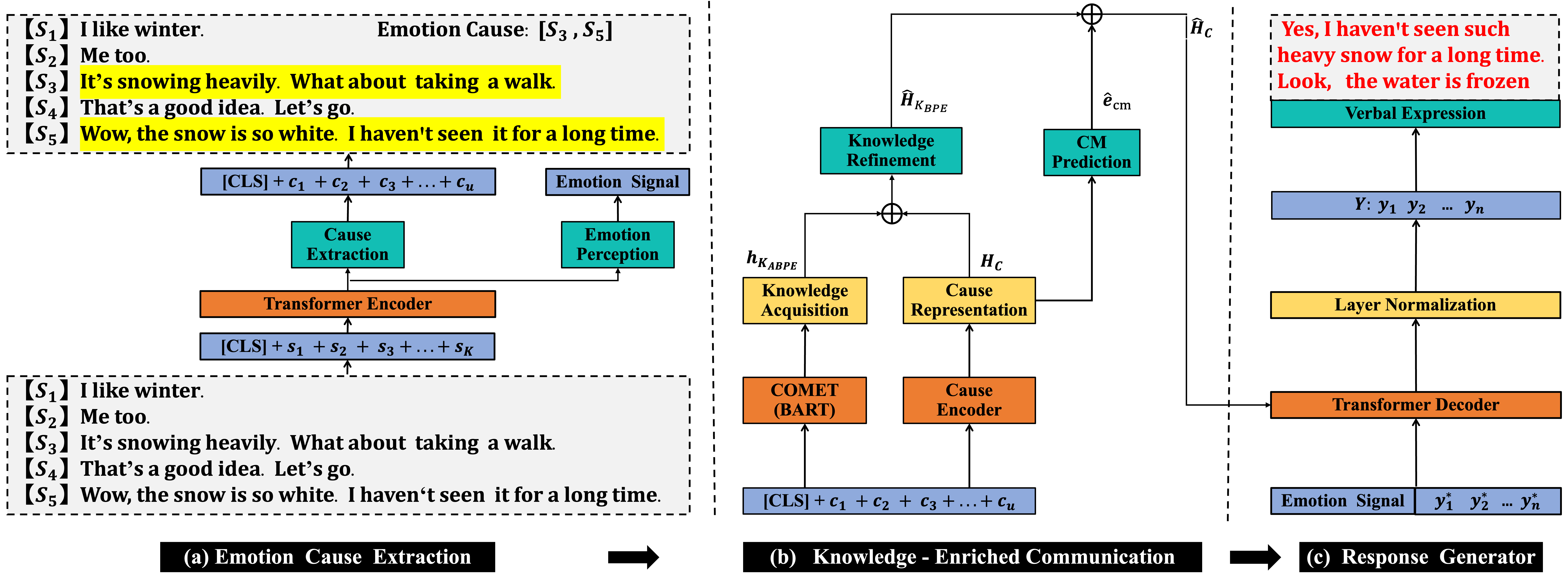}} 
\caption{An overall framework of \textbf{IMAGINE}.} 
\label{sec:IMAGINE}
\end{figure*}

\subsection{What affects empathy? }
\noindent \textbf{Emotion Cause}  \qquad  The emotion cause (also called antecedents, triggers, or
stimuli) \citep{ellsworth2003} is a stimulus for human emotions. 
Recognizing the emotion cause helps understand human emotions better to generate more empathetic responses. 
The cause could also be a speaker’s counterpart reacting towards an event cared for by the speaker(inter-personal emotional influence).
For example, understanding the sentence, "I like summer as it is a great time to surf," is not only to detect the positive emotion, HAPPY, but also to find its cause -- "it is a great time to surf."
The emotion cause recognition method \citep{poria2021} is used in our work.

\noindent \textbf{External Knowledge}  \qquad  A major part of cognitive empathy is understanding the situations and feelings of others.
Conversations are limited in time and content.
Therefore, using our experience (e.g., external knowledge) is important to connect what is explicitly mentioned and what is associated with it. 
In this work, we use the ATOMIC-2020 dataset \citep{hwang2020} as our commonsense knowledge base, which is a collection of commonsense reasoning inferences about everyday if-then contexts. 
Detailed information about ATOMIC is covered in Appendix ${\color{red}{\fbox{\ref{sec:appendix}}}}$. 



\noindent \textbf{Communication Mechanism (CM)}  \qquad For empathy generation, both conveying cognitive understanding \citep{truax1967} and expressing stimulated emotions \citep{davis1980} are essential. 
\citet{sharma2020} presented a computational approach to understanding empathy expressed in textual, asynchronous conversations and addressing both emotional and cognitive aspects of empathy. 
They developed components of an empathetic expression, consisting of three communication mechanisms - $\textbf{Emotional Reaction}$ (expressing emotions such as warmth, compassion, and concern), $\textbf{Interpretation}$ (conveying an understanding of feelings and experiences), and $\textbf{Exploration}$ (improving understanding of the seeker by exploring the feelings and experiences).

\subsection{Task Formulation}
We formulate the task of empathetic response generation as follows.
Given dialogue transcripts \textbf{S} = $\{\mathbf{s}_0, \mathbf{s}_1, ... ,\mathbf{s}_k\}$ with $k$ utterances,
we firstly detect the emotion and extract emotion causes \textbf{C} = $\{\mathbf{c}_0, \mathbf{c}_1, ... ,\mathbf{c}_{u}\}$ which are a subset of $\textbf{S}$.
Each utterance $\mathbf{c}_i$ = $\{\mathbf{c}_{i,1}, \mathbf{c}_{i,2}, ... ,\mathbf{c}_{i,l_{i}}\}$ is a sequence of tokens, where $\mathbf {}l_{i}$ denotes the length. 
Then, our goal is to generate an empathetic response \textbf{Y} = $\{\mathbf{y}_1, \mathbf{y}_2, ... ,\mathbf{y}_{n}\}$ given the sequence \textbf{C}, with the assistance of external knowledge and communication mechanisms.

\section{Approach}
Our proposed model, \textbf{IMAGINE}, is built upon the standard Transformer \citep{vaswani2017} and its overview is illustrated in Figure ${\color{red}{\fbox{\ref{sec:IMAGINE}}}}$. 
It has three stages consisting of five functional modules: emotion cause extraction (emotion perception, cause extraction), knowledge-enriched communication (dialogue  reaction, experience projection), and response generation (verbal expression).
Emotion perception predicts emotions of the input.
Cause extraction extracts causes related to the emotions from the input.
Experience projection acquires knowledge based on the causes mentioned above. 
Dialogue reaction decides the response strategies by learning from the contexts.
Verbal expression integrates the information obtained from the above four modules and generates appropriate responses. 

\subsection{Emotion Cause Extraction}
Given a dialogue context consisting of $k$ utterances with the context emotion, the goal of emotion cause extraction is to identify which utterances in the dialogue context contain the emotion cause.
We leverage an existing model which is trained on an open-domain emotional dialogue dataset named RECCON, for identifying emotion causes at utterance level in conversations \citep{poria2021}. 
\citet{gao2021} has verified the model's validity, and we follow the method in the first stage of our work. 

\subsubsection{Emotion Perception} 

It is a classification problem aiming
at predicting the emotion $\bm{\varepsilon}$ within the dialogue context.
Given the dialogue context \textbf{S} = $\{\mathbf{s}_0, \mathbf{s}_1, ... ,\mathbf{s}_k\}$ as the input, the tokens are then fed into a transformer-based encoder to obtain
a sequence of contextualized representations  $\mathbf{H}_{S}$.
Hence, we pass $\mathbf{H}_{{S}}$ through a linear layer followed by a softmax operation to produce the emotion category distribution:
\begin{equation}
 \mathbf{\hat{e}}_{emo} = \mathbf{W}_{e}\mathbf{H}_{{S}}[0] \mathbf{{}+} \mathbf{b}_{e}\label{emo:inventoryflow},
\end{equation}
\begin{equation}
 \hat{\mathbf{P}}({\bm{\varepsilon}|{\mathbf{S}}}) = \mathbf{softmax}(\mathbf{\hat{e}}_{emo}),
\end{equation}
where $\mathbf{W}_{e}$ and $\mathbf{b}_{e}$ are trainable parameters. 
During training, we employ negative log-likelihood as the emotion perception loss:
\begin{equation}
 \mathbf{{}L}_{emo} = \mathbf{{}{-log}(\hat{\mathbf{P}}({\bm{\varepsilon} = \mathbf{e^*}  |{\bm{S}}}))},
\end{equation}
where $\mathbf{e}^{*}$ denotes the emotion label, and $\bm{\varepsilon}$ denotes the predicted output. 
Emotional vectors $\mathbf{\hat{e}_{emo}}$ will be fed into the decoder as a crucial emotional signal to
guide the empathetic response generation.

\subsubsection{Cause Extraction} 

Given the dialogue context \textbf{S} and its emotion $\bm{\varepsilon}$, we extract emotion causes \textbf{C} = $\{\mathbf{c}_0, \mathbf{c}_1, ... ,\mathbf{c}_{u}\}$ according to the approach in \citet{poria2021}.  
The causes \textbf{C} are a subset of \textbf{S} and will be used as the input of the next two stages.
Following previous work \cite{lin2019,majumder2020,sabour2021}, we concatenate the utterances indicating emotion causes and prepend a special token $\mathbf[CLS]$ to obtain the cause input \textbf{C} = $\mathbf[CLS]$ + $\mathbf{c}_0$ + $\mathbf{c}_1$ + $\mathbf...$ + $\mathbf{c}_{u}$.  
Each utterance  $\mathbf{c}_i$ contains a sequence of tokens:  $\mathbf{c}_i$ = $\{\mathbf{c}_{i,1}, \mathbf{c}_{i,2}, ... ,\mathbf{c}_{i,l_{i}}\}$, where $\mathbf {}l_{i}$ is the length of $\mathbf{c}_i$.

Each token is represented from three aspects: its semantic meaning, its position in the sequence, and who said it.
Suppose that the token ID and the position ID of $\mathbf{c}_{i,j}$ are ${w}_{{\bm{c}}_{i,j}}$ $\in$ [0, |$\mathbf{V}$|) ($\mathbf{V}$ is the vocabulary) and ${p}_{{\bm{c}}_{i,j}}$, respectively.
Additionally, in multi-turn dialogue settings, distinguishing a listener from a speaker is helpful. 
So we incorporate the dialogue state embedding into our input sequence.
Specifically, each utterance $\mathbf{c}_i$ is labeled with its corresponding role ${s}_{\bm{c}_{i}}$ $\in$ $\{$0, 1$\}$ (0 for speaker and 1 for listener). 

The token $\mathbf{c}_{i,j}$ is represented by summing up the word embedding, positional embedding, and dialogue state embedding:
\begin{equation}
 \mathbf{E}_{\bm{c}_{i,j}} = \mathbf{E}_{W}[w_{\bm{c}_{i,j}}] + \mathbf{E}_{P}[p_{\bm{c}_{i,j}}] + \mathbf{E}_{S}[s_{\bm{c}_{i}}],
\end{equation}
where $\mathbf{E}_{W}$ $\in$ $\mathbb{R}^{|\mathbf{}{V}| \times \mathbf{}{d}}$, $\mathbf{E}_{P}$ $\in$ $\mathbb{R}^{\mathbf{}{1024} \times \mathbf{}{d}}$, $\mathbf{E}_{S}$ $\in$ $\mathbb{R}^{\mathbf{}{2} \times \mathbf{}{d}}$ denote the embedding matrices of word, position, and state.
[·] denotes the indexing operation, and $\mathbf{}{d}$ is the dimensionality of embeddings.
We feed the entire sequence of token embeddings $\mathbf{E}_C$ organized by $\mathbf{E}_{\bm{c}_{i,j}}$ to a cause encoder to produce the contextual representation:
\begin{equation}
 \mathbf{H}_C = \mathbf{Cause\textbf{-}Encoder}(\mathbf{E}_C),
\end{equation}
where $\mathbf{H}_{C}$ $\in$ $\mathbb{R}^{|\mathbf{}{L}| \times \mathbf{}{d}}$, $\mathbf{}{L}$ is the length of the sequence, and $\mathbf{}{d}$ is the hidden size of the cause encoder.

Next, we use the hidden state at the $\mathbf[CLS]$ of the cause encoder, $\mathbf{h}_{c}$ = $\mathbf{H}_{C}[0]$, to predict CM  strategies in the following stage.


\subsection{Knowledge - Enriched Communication}

\subsubsection{Dialogue Reaction}
\noindent \textbf{CM Prediction}  \qquad While no empathetic conversation corpora provide annotations of diverse empathy factors, there are abundant publicly available resources that make automatic annotation feasible. 
We use two corpora annotated with CM provided by \citet{sharma2020}.
There are three communication factors named Emotion Reaction (ER), Interpretation (IP), and Exploration (EX).
Each mechanism has different degrees.
In our work, we merge "weak" and "strong" into "yes" and differentiate each mechanism's degree into two types: "no" and "yes". 
 
We pass $\mathbf{h}_{c}$ through a linear layer followed by a softmax operation to produce the CM category distribution:
\begin{equation}
 \mathbf{e}_{cmi} = \mathbf{W}_{cmi}\mathbf{h}_{c} \mathbf{{}+} \mathbf{b}_{cmi}, \quad cmi \in \left\{er,ip,ex\right\} 
\end{equation}
\begin{equation}
 \hat{\mathbf{P}}_{cmi} = \mathbf{softmax}(\mathbf{e}_{cmi}),
\end{equation}
The negative log-likelihood loss is calculated:
\begin{equation}
 \mathbf{{}L}_{cm} = \sum_{cmi \in \left\{er,ip,ex\right\}}  \mathbf{-log}(\hat{\mathbf{P}}_{cmi}),
\end{equation}
Finally,  $\mathbf{e}_{er}$, $\mathbf{e}_{ip}$, $\mathbf{e}_{ex}$ are summed up, weighted by their predicted degree, as a crucial CM signal:
\begin{equation}
 \mathbf{\hat{e}}_{cm} = \hat{\mathbf{P}}_{er} \cdot \mathbf{e}_{er} + \hat{\mathbf{P}}_{ip} \cdot \mathbf{e}_{ip} + \hat{\mathbf{P}}_{ex} \cdot \mathbf{e}_{ex}\label{con:inventoryflow},
\end{equation}

\subsubsection{Experience Projection}

\begin{figure}[t!] 
\center{\includegraphics[width=6cm]  {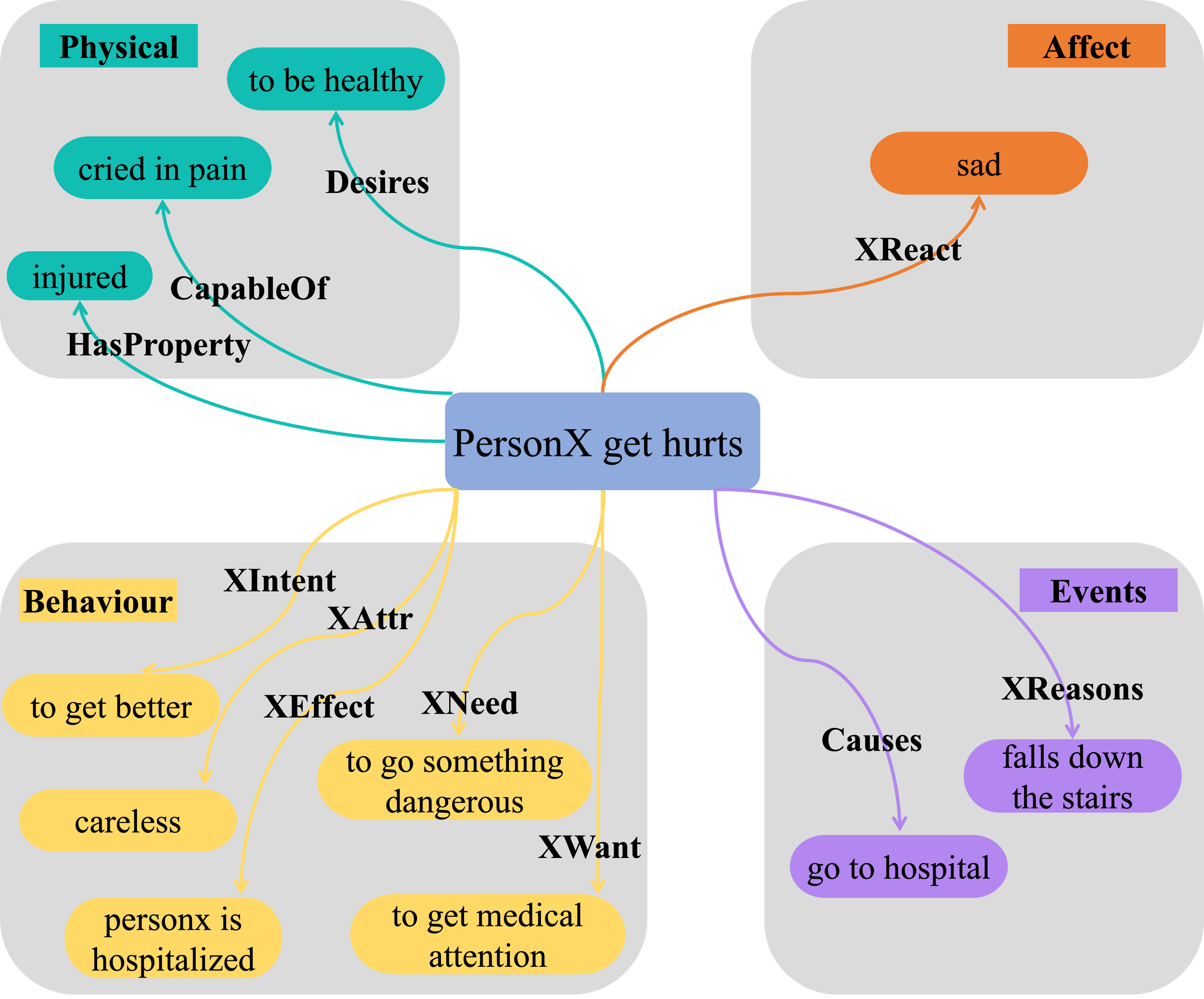}} 
\caption{The four modules of the Knowledge Graph.} 
\label{sec:knowledge}
\end{figure}

\noindent \textbf{Knowledge Acquisition} \qquad  We extend the contexts by selecting from the knowledge graph those that are speaker-centered and contribute positively to the speaker.
Finally, we split ATOMIC-2020 \citep{hwang2020} into four types: \textit{Affect}, \textit{Behaviour}, \textit{Physical}, and \textit{Events}, containing 11 relations $[r_1, r_2, ..., r_{11}]$ in total (See Figure ${\color{red}{\fbox{\ref{sec:knowledge}}}}$). 
In Affect, we select one relation: ($\mathbf[XReact]$). 
In Behaviour, we select five relations: ($\mathbf[XIntent]$, $\mathbf[XNeed]$, $\mathbf[XWant]$, $\mathbf[XEffect]$, $\mathbf[XAttr]$). 
In Physical, we select three relations: ($\mathbf[HasProperty]$, $\mathbf[CapableOf]$, $\mathbf[Desires]$). 
In Events, we select two relations: ($\mathbf[Causes]$, $\mathbf[XReason]$). 
For an input sequence \textbf{C}, we use \textbf{COMET} \citep{lewis2019} to generate five commonsense-inferred entities  [$\mathbf{}{s}^{r_i}_1$, $\mathbf{}{s}^{r_i}_2$, $\mathbf{}{s}^{r_i}_3$, $\mathbf{}{s}^{r_i}_4$, $\mathbf{}{s}^{r_i}_5$] for each relation $r_i$. 
Then we concatenate all entities generated from relations belonging to the same relation type.
Through this way, we obtain four commonsense sequences for each input sequence: $\mathbf{}{S}_{Affect}$, $\mathbf{}{S}_{Behav}$, $\mathbf{}{S}_{Phys}$, and $\mathbf{}{S}_{Events}$. 
For example, $\mathbf{}{S}_{Events}=[\mathbf{}{s}^{[Causes]}_1, ..., \mathbf{}{s}^{[Causes]}_5, \mathbf{}{s}^{[XReasons]}_1, ..., \mathbf{}{s}^{[XReasons]}_5]$.
We prepend $\mathbf[CLS]$ to $\mathbf{}{S}_{Behav}$, $\mathbf{}{S}_{Phys}$, and $\mathbf{}{S}_{Events}$.
$\mathbf{}{S}_{Affect}$ does not change because the entities for \textit{Affect} are usually independent emotion words (e.g., happy, surprise, sad) rather than semantically coherent sequences. 
The commonsense sequences are fed to the knowledge encoder:
\begin{equation}
 \mathbf{H}_{K_{ABPE}} = \mathbf{Knowledge\textbf{-}Encoder}(\mathbf{\textsl{S}}_{K_{ABPE}}),
\end{equation}
where $K_{ABPE} \in \left\{Affect,Behav,Phys,Events\right\}$.
$\mathbf{H}_{K_{ABPE}}$$\in$$\mathbb{R}^{|\mathbf{}{L}_{K_{ABPE}}| \times \mathbf{}{d}}$, with $\mathbf{}{|\mathbf{}{L}_{K_{ABPE}}|}$ being lengths of the commonsense entity sequences.

Next, we use hidden representations of the first position to represent sequences  $\mathbf{}{S}_{Behav}$, $\mathbf{}{S}_{Phys}$, and $\mathbf{}{S}_{Events}$, respectively:
\begin{equation}
\mathbf{h}_{K_{BPE}} = \mathbf{H}_{K_{BPE}}[0]
\end{equation}
where $K_{BPE} \in \left\{Behav,Phys,Events\right\}$.

Moreover, we use the mean of hidden representations to represent $\mathbf{}{S}_{Affect}$:
\begin{equation}
 \mathbf{h}_{Affect} = \mathbf{Average}(\mathbf{H}_{Affect})|_{axis=0},
\end{equation}

\noindent \textbf{Knowledge Refinement} \qquad  In order to refine the emotion causes by knowledge information, we concatenate each commonsense relation representation $\mathbf{h}_{K_{ABPE}}$ to the cause representation $\mathbf{H}_{C}$ at the token level.
In contrast to sequence-level concatenation, token-level concatenation enables us to fuse knowledge within each word in the cause sequence:
\begin{equation}
 \mathbf{U}_{K_{ABPE}} = \mathbf{H}_{C} \oplus \mathbf{h}_{K_{ABPE}},
\end{equation}
where $\mathbf{U}_{Affect}$,$\mathbf{U}_{Behav}$,$\mathbf{U}_{Phys}$,$\mathbf{U}_{Events}$$\in$$\mathbb{R}^{|\mathbf{}{L}| \times \mathbf{}{2d}}$.

Accordingly, we encode the fused representations and obtain knowledge-refined cause representations for each relation type:
\begin{equation}
 \mathbf{H}^{ref}_{K_{ABPE}} = \mathbf{Refine\textbf{-}Encoder}({\mathbf{{U}}_{K_{ABPE}}}),
\end{equation}
where $\mathbf{H}^{ref}_{K_{Affect}}$,$\mathbf{H}^{ref}_{K_{Behav}}$,$\mathbf{H}^{ref}_{K_{Phys}}$,$\mathbf{H}^{ref}_{K_{Events}}$ $\in$$\mathbb{R}^{|\mathbf{}{L}| \times \mathbf{}{d}}$.

We believe that relations of the \textit{Affect} type matter to emotional empathy, meanwhile relations of \textit{Behavior}, \textit{Physical},  and \textit{Events} types matter to cognitive empathy.
Hence, we re-represent the knowledge-refined cause representations as below:

\begin{equation}
 \mathbf{\tilde{H}}_{K_{BPE}} = \mathbf{H}^{ref}_{K_{BPE}} \oplus \mathbf{H}^{ref}_{Affect},
\end{equation}
where $\mathbf{\tilde{H}}_{Behav}$,$\mathbf{\tilde{H}}_{Phys}$,$\mathbf{\tilde{H}}_{Events}$$\in$$\mathbb{R}^{|\mathbf{}{L}| \times \mathbf{}{2d}}$.

Next, to highlight important features within the knowledge-refined cause representation, we assign importance scores to $\mathbf{\tilde{H}}_{K_{BPE}}$, followed by a Multi-Layer Perception (MLP) layer with ReLU:
\begin{equation}
 \mathbf{\hat{H}}_{K_{BPE}} = \mathbf{MLP}{(\sigma(\tilde{\mathbf{H}}}_{K_{BPE}}) \cdot \tilde{\mathbf{H}}_{K_{BPE}})
\end{equation}
where $\mathbf{\hat{H}}_{Behav}$,$\mathbf{\hat{H}}_{Phys}$,$\mathbf{\hat{H}}_{Events}$$\in$$\mathbb{R}^{|\mathbf{}{L}| \times \mathbf{}{d}}$, and $\cdot$ denotes element-wise multiplication.

Finally, $\mathbf{\hat{H}}_{Behav}$,$\mathbf{\hat{H}}_{Phys}$,$\mathbf{\hat{H}}_{Events}$ and $\hat{\mathbf{e}}_{cm}$\\(Equation ${\color{red}{\fbox{\ref{con:inventoryflow}}}}$), are fed into the decoder:
\begin{equation}
 \mathbf{\hat{H}}_{C} = \mathbf{\hat{H}}_{Behav} \oplus \mathbf{\hat{H}}_{Phys} \oplus \mathbf{\hat{H}}_{Events} \oplus \mathbf{\hat{e}}_{cm}
\end{equation}
where $\mathbf{\hat{H}}_{C}$$\in$$\mathbb{R}^{|\mathbf{}{L}| \times \mathbf{}{4d}}$.

\subsection{Response Generation}
\noindent \textbf{Verbal Expression}  \qquad To acquire emotion dependencies, we concatenate the intermediate emotional signal $\hat{\mathbf{e}}_{emo}$ with word embeddings of the expected response and get [$\mathbf{y}_{0}^*$,$\mathbf{y}_{1}^*$, $\mathbf{y}_{2}^*$, $\bm{...}$, $\mathbf{y}_{n}^*$].
Here $\mathbf{y}_{0}^*$ is $\hat{\mathbf{e}}_{emo}$.
We then feed the embeddings into the response decoder. Our decoder is built based on Transformer layers:
\begin{equation}
 \mathbf{P}({{y}_{t}|{y_{<t},\mathbf{C}}}) = \mathbf{Decoder}(\mathbf{E}_{y_{<t}},\mathbf{\hat{H}}_{C}),
\end{equation}
where $\mathbf{E}_{y_{<t}}$ denotes embeddings of tokens that have been generated. 
Note that the cross attention to the encoder outputs is modified to the knowledge-refined cause representation $\mathbf{\hat{H}}_{C}$, which has fused the information from both the cause and the commonsense-inferred entities.

\subsection{Model Training}
We use negative log-likelihood of the ground-truth words $\mathbf{y}^*_{t}$ as the generation loss function:
\begin{equation}
\mathbf{}{\bm{L}}_{gen} = -\sum^n_{t=1} \mathbf{log}\mathbf{P}(\mathbf{y}_t = \mathbf{y}_t^* | \mathbf{y}_0, ... ,\mathbf{y}_{t-1},\mathbf{C })
\end{equation}

Dialogue generation models sometimes generate repetitive phrases or generic responses, such as "That is a good idea" and "Oh, it is bad." 
To solve this problem, we apply the Response Diversity Loss in our model, implementing Frequency-Aware Cross-Entropy (FACE) \citep{Jiang2019} as an additional loss to penalize high-frequency tokens using a weighting
scheme. 
Hence, during training, prior to receiving a new batch of samples, we derive the frequency-based weight $\mathbf{w}_{i}$ for each vocabulary token $\mathbf{v}_{i}$ in the training corpus:
\begin{equation}
 \mathbf{w}_{i} = \mathbf{a} \times\mathbf{FQ}_i + 1 ,
\end{equation}
\begin{equation}
 \mathbf{FQ}_{i} = \frac{\mathbf{freq}(\mathbf{v}_{i})}{\sum^V_{j=1} \mathbf{freq}(\mathbf{v}_{j})} ,
\end{equation}
where V denotes the vocabulary size, $\mathbf{a} = - (\mathbf{max}_{0<j<V} (\mathbf{FQ}_{j}))^{-1}$ is the frequency slope and 1 is added as the bias so that $\mathbf{w}_{i}$ falls into [0,1]. 
Lastly, we normalize  $\mathbf{w}_{i}$ to have mean of 1, as done by \citep{Jiang2019}.
The diversity loss would then be calculated as below:
\begin{equation}
\mathbf{}{\bm{L}}_{div} = -\sum^n_t\sum^V_i \mathbf{w}_{i} \delta(\mathbf{v}_{i}=\mathbf{y}_{t}^*) \mathbf{log} \mathbf{P}(\mathbf{v}_{i} | \mathbf{y}_{<t}, \mathbf{C})
\end{equation}
where $\mathbf{v}_{i}$ is a candidate token in the vocabulary and $\delta$
is the indicator function, which equals to 1 if and only if $\mathbf{v}_{i}$ = $\mathbf{y}_{t}^*$ and 0 otherwise. All parameters of our proposed model are trained and optimized based on the weighted sum of four losses:
\begin{equation}
\mathbf{}{\bm{L}} = \lambda_1 \mathbf{}{\bm{L}}_{gen} + \lambda_2 \mathbf{}{\bm{L}}_{emo} + \lambda_3 \mathbf{}{\bm{L}}_{cm} + \lambda_4 \mathbf{}{\bm{L}}_{div},
\end{equation}
where $\lambda_1$, $\lambda_2$, $\lambda_3$ and $\lambda_4$ are hyper-parameters that we use to control the influence of the four losses. Loss weights $\lambda_1$, $\lambda_2$, $\lambda_3$ and $\lambda_4$ are set to 1, 1, 1, and 1.5, respectively.

\section{Experimental Settings}
\subsection{Dataset}
We conduct our experiments on the EMPATHETICDIALOGUES dataset \citep{rashkin2018}.
It is a large-scale multi-turn empathetic dialogue dataset containing 25k dialogue sessions, each having 3-5 rounds of dialogue.
There are 32 different distributions of emotion labels. 
Following the original dataset definitions, we use the 8:1:1 train/valid/test subset split.

\subsection{Comparison Methods}
The following models are selected as baselines:
\noindent \textbf{1) Transformer} \citep{vaswani2017}:  A Transformer based encoder-decoder model.

\noindent \textbf{2) Multi-TRS} \citep{rashkin2018}:  An extension of the Transformer model that has an additional unit for emotion prediction.

\noindent \textbf{3) MoEL} \citep{lin2019}:  Another extension of Transformer model which softly combines
the response representations from different decoders.

\noindent \textbf{4) MIME} \citep{majumder2020}:  Another extension of transformer model which considers emotion
clustering and emotional mimicry. Besides, it also introduces sampling stochasticity during training.

\noindent \textbf{5) EMPDG} \citep{li2019}:  A multi-resolution empathetic adversarial chatbot which exploits multi-resolution emotions and user feedback.

\noindent \textbf{6) CEM} \citep{sabour2021}:  A Transformer encoder-decoder model that integrates affection and cognition into commonsense knowledge.

\noindent \textbf{7) KEMP} \citep{li2022}:  A contextual-enhanced empathetic dialogue generator that leverages multi-type external knowledge and emotional signal distilling for response generation.

More implementation details of our \textbf{IMAGINE} model is covered in Appendix ${\color{red}{\fbox{\ref{sec:expriments}}}}$.

\begin{table*}[t]
    \centering
    \resizebox{\textwidth}{!}
    {
    \begin{tabular}{lccccrrrr}
    \toprule
        \textbf{Models} & \textbf{PPL} & \textbf{BLEU-2} & \textbf{Dinstinct-1} & \textbf{Distinct-2} & \textbf{ACC} & \textbf{Fluency} & \textbf{Relevance} & \textbf{Empathy} \\ \midrule
        Transformer & 37.62 & 1.32 & 0.45 & 2.02 & - & 3.04 & 2.49 & 2.50\\ 
        Multi-TRS & 37.75 & 1.31 & 0.41 & 1.67 & 33.57 & 2.99 & 2.51 & 2.59\\
        MoEL & 36.93 & 1.32 & 0.44 & 2.10 & 30.62 & 3.28 & 2.57 & 2.63\\
        MIME & 37.09 & 1.34 & 0.47 & 1.90 & 31.36 & 3.14 & 2.52 & 2.59\\
        EmpDG & 37.29 & 1.30 & 0.46 & 2.02 & 30.41 & 3.07 & 2.69 & 2.72\\
        CEM & 36.11 & 1.35 & 0.66 & 2.99 & 39.11 & 3.40 & 2.96 & 2.94\\
        KEMP & 36.89 & 1.34 & 0.55 & 2.29 & 39.31 & 3.27 & 2.68 & 2.68\\
        \hline
        \textbf{IMAGINE} & \textbf{35.10} & \textbf{1.37} & \textbf{0.76} & \textbf{3.40} & \textbf{39.60} & \textbf{3.58} & \textbf{3.09} & \textbf{3.09}\\
    \bottomrule
    \end{tabular}
    }
    \caption{Results of automatic and human evaluations.}
    \label{tab:1}
\end{table*}

 \begin{table*}[t]
     \centering
    \begin{tabular}{lcccrr}
    \toprule
        \textbf{Models} & \textbf{PPL} & \textbf{BLEU-2} & \textbf{Dinstinct-1} & \textbf{Distinct-2} & \textbf{ACC}  \\ \midrule
        \textbf{IMAGINE} & 35.10 & \textbf{1.37} & \textbf{0.76} & \textbf{3.40} & \textbf{39.60}\\
        W/O cause & 35.43 & 1.35 & 0.64 & 2.57 & 38.60 \\ 
        W/O cm & 35.58 & 1.34 & 0.63 & 2.84 & 38.88 \\ 
        W/O know & 35.00 & 1.36 & 0.64 & 2.92 & 38.50 \\ 
        W/O DIV & \textbf{34.50} & 1.37 & 0.68 & 2.94 & 39.10 \\ 
    \bottomrule
    \end{tabular}
    \caption{Ablation study.}
    \label{tab:2}
\end{table*}

\begin{table}[t]
    \centering
    \begin{tabular}{lccr}
    \toprule
        \textbf{Models} & \textbf{Win}\bm{$\%$} & \textbf{Lose}\bm{$\%$} & \textbf{Tie}\bm{$\%$} \\ \midrule
        Ours VS Transformer & 49.18 & 16.83 & 33.99\\
        Ours VS Multi-TRS & 42.34 & 17.66 & 40.00 \\
        Ours VS MOEL & 45.49 & 27.42 & 27.09 \\
        Ours VS MIME & 47.34 & 19.33 & 33.33 \\
        Ours VS EmpDG & 47.18 & 19.60 & 33.22 \\
        Ours VS CEM & 42.96 & 25.80 & 31.24 \\
        Ours VS KEMP & 41.90 & 23.98 & 34.12\\
        
    \bottomrule
    \end{tabular}
    \caption{Results of human A/B test.}
    \label{tab:3}
\end{table}

\subsection{Evaluation metrics}
\noindent \textbf{Automatic Evaluations}  \qquad Four automatic metrics are applied for evaluation:

\noindent \textbf{1) PPL} \citep{serban2015}:  The perplexity (PPL) represents the model’s confidence in its set of candidate responses.
A low PPL value means high confidence.
PPL can be used to evaluate the general quality of the generated responses.

\noindent \textbf{2) BLEU-2} \citep{papineni2002}: It calculates the co-occurrence frequency of n-grams between candidates and references.

\noindent \textbf{3) Distinct-1 and Distinct-2} \citep{li2015}:  It is the proportion of the distinct unigrams/bigrams in all the generated results to indicate the diversity.

\noindent \textbf{4) ACC}:  To evaluate the model at the emotional level, we adopt Emotion Accuracy (ACC) as the agreement between the ground truth emotion labels and the predicted emotion labels.

\noindent \textbf{Human Ratings}  \qquad Evaluating open-domain dialogue systems is challenging due to the lack of reliable automatic evaluation metrics \citep{gao2021b}. 
Thus, human judgments are necessary. 
We randomly sample 100 dialogues and generate corresponding responses from different models. 
Five well-educated native English speakers who work in literary writing, psychology, and teaching are hired to give each response a rating score from three aspects -- Fluency, Relevance, and Empathy. 
Each aspect is on a scale from 1 to 5, where 1, 2, 3, 4, and 5 indicate unacceptable, not good, moderate, good, and excellent performance, respectively.
In order to keep the anonymization of compared methods, the order of responses in each dialogue is shuffled.

\noindent \textbf{Human A/B Test} \qquad In the human A/B test, to make sure fairness, we re-sample another 700 dialogues (100 for each comparison between our model and a baseline model) and form them into A-vs-B types, where A is our model and B is a baseline model. 
Another three annotators are asked to choose a better response.
They can also choose a Tie if they think both are good or bad.
All human evaluation tasks are conducted on \url{https://www.fanhantech.com}.

\begin{figure*}[htbp] 
 \center{\includegraphics[width=14cm]  {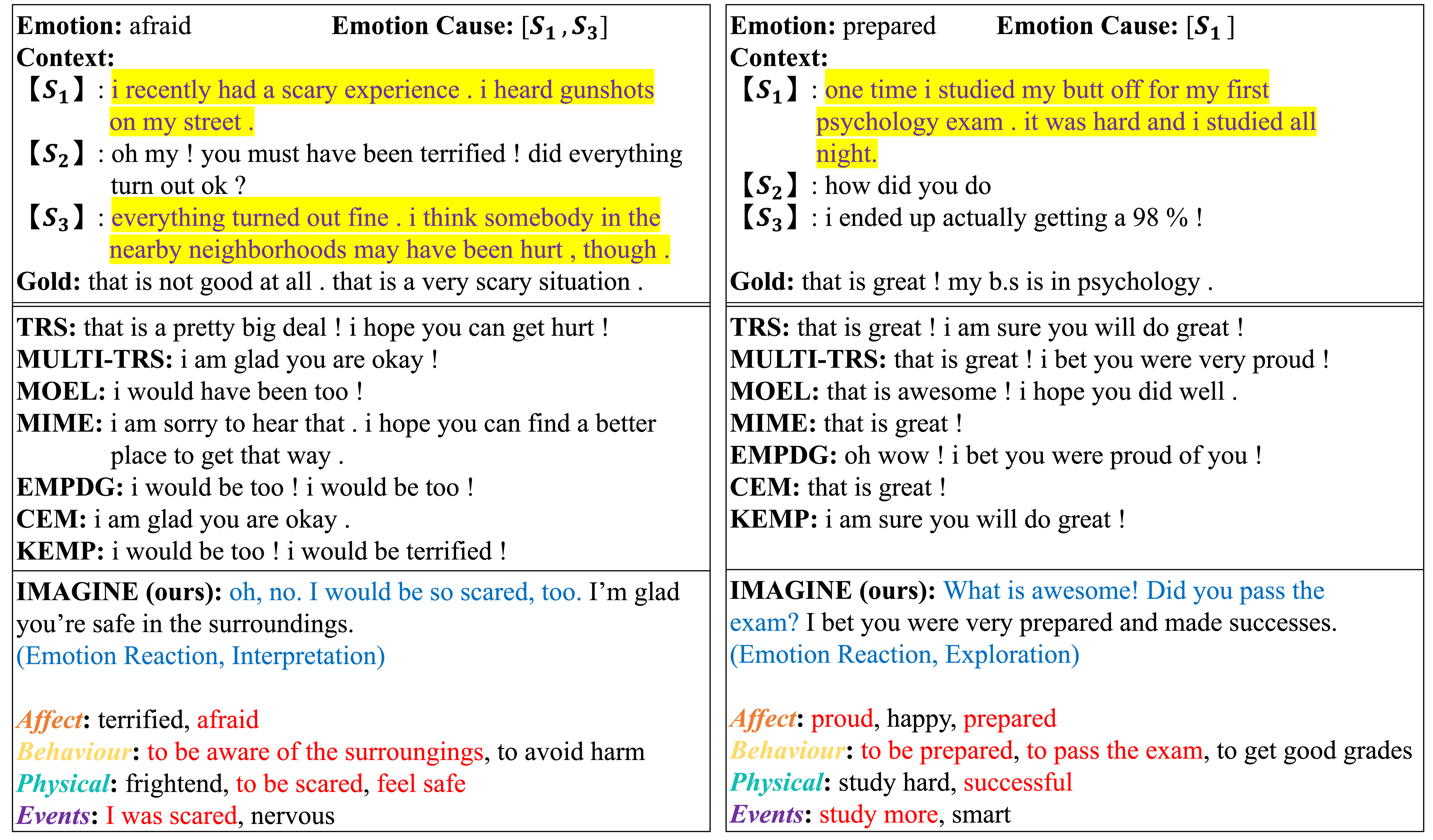}} 
 \caption{Case study of the generated responses by \textbf{IMAGINE} and the baselines.} 
 \label{sec:case}
 \end{figure*}

\section{Experimental Results}
\subsection{Automatic Evaluation Results}
Table ${\color{red}{\fbox{\ref{tab:1}}}}$ reports the evaluation results on automatic metrics. Our model \textbf{IMAGINE} achieves the lowest perplexity, indicating that the overall quality of our generated responses is higher than the baselines. 
Moreover, the results of Distinct-1 and Distinct-2 show that our model generates much more diverse responses than baselines. 
As for the emotion accuracy, we can see that our model is valid for recognizing emotions.

\subsection{Human Evaluation Results}
Table ${\color{red}{\fbox{\ref{tab:1}}}}$ illustrates that \textbf{IMAGINE} obtains the best performance on Fluency, Relevance, and Empathy scores. 
It proves that integrating emotion causes, knowledge, and communication mechanisms can generate more informative and empathetic responses.
In addition, from the results of the human A/B test in Table ${\color{red}{\fbox{\ref{tab:3}}}}$, we see that responses from \textbf{IMAGINE} are more often preferable to humans than the responses from other baseline models, which strongly supports the advantages of our approach.
 
\subsection{Ablation Analysis}
We conducted ablation studies to verify the effectiveness of each component in our model. 
Table ${\color{red}{\fbox{\ref{tab:2}}}}$ reports the results. 

\noindent \textbf{1) W/O cause}: 
Looking at Table ${\color{red}{\fbox{\ref{tab:2}}}}$, we can see that removing the emotion cause extraction part leads to a significant performance decrease of both models in terms of response generation and emotion recognition. 
The original dialogue history may contain emotion-irrelevant information, which results in a shift of focus.
The result indicates that emotion cause extraction plays an important role in strengthening the understanding of users' emotions, which improves the generation of empathetic responses.

\noindent \textbf{2) W/O CM}: By removing the communication mechanism from the response generation module, as shown in Table ${\color{red}{\fbox{\ref{tab:2}}}}$, we can see that our model is less empathetic and also has a tendency to decline in emotion prediction. 
The communication mechanism is a state of understanding how people feel; without it, our model will have fewer communication skills.

\noindent \textbf{3) W/O know}: When we remove the knowledge module,  as shown in Table ${\color{red}{\fbox{\ref{tab:2}}}}$, we can see that the quality and diversity of the model's responses are declined, as a lack of knowledge leads to weaker ability to enrich emotion causes.
It also affects the closeness and relevance of the generated responses to the context.

\noindent \textbf{4) W/O DIV}: If the diversity loss is removed, we can see from Table ${\color{red}{\fbox{\ref{tab:2}}}}$ that Distinct-1 is reduced from 0.76 to 0.68, and Distinct-2 is reduced from 3.4 to 2.94.
It indicates the effectiveness of this loss in generating more diverse responses.
 
\subsection{Case Study}
We also present some examples of responses generated by our models and baseline models in Figure ${\color{red}{\fbox{\ref{sec:case}}}}$.
Compared with baseline models, our model generates responses closer to the "gold" responses.
As shown in the first example, our model can reason deeply about the emotion cause and get a good result in terms of knowledge acquisition.
In the second example, from the dialogue context, we learn that the user "studied hard and got good grades."
Through the knowledge base, we infer richer information like "prepared, successful, and pass the exam."
Finally, our model congratulates and praises the user and poses an unasked question to him/her.

\section{Conclusion}
This paper presents a novel framework that integrates emotion causes, knowledge graphs, and communication mechanisms for empathetic response generation. 
The emotion cause detection allows us to determine what events stimulate a user's emotion. 
We can understand the events with the knowledge graph, enriching the contextual information. 
Furthermore, the communication mechanisms enhance our ability to let users feel that we are tryining to feel what they feel. 
Automatic and human evaluations show that our proposed approach can generate more informative and empathetic responses.

\section*{Limitations}
The first challenge is a common problem current chatbots face, e.g., traceability of models and reasoning ability.
Second, for mental health support chatbots, each person is analyzed on a case-by-case basis.
Each person with a mental health impairment needs a personalized approach to communication, which is not overly generalized.
Finally, the shortcomings of the knowledge graph - size, breadth, diversity, and rationality - directly determine the quality of the causes' associative expansion and also affect the closeness and relevance of the generated responses to the context.


\section*{Ethics Statement}
The empathetic-dialogues dataset \citep{rashkin2018} used in our paper protects the privacy of real users. 
Furthermore, we make sure anonymization in the human evaluation process. 
We believe our research work meets the ethics of EMNLP.


\bibliography{emnlp2022}
\bibliographystyle{acl_natbib}

\appendix

\section{knowledge Graph}
\label{sec:appendix}
\begin{figure*}[htbp] 
 \center{\includegraphics[width=16cm]  {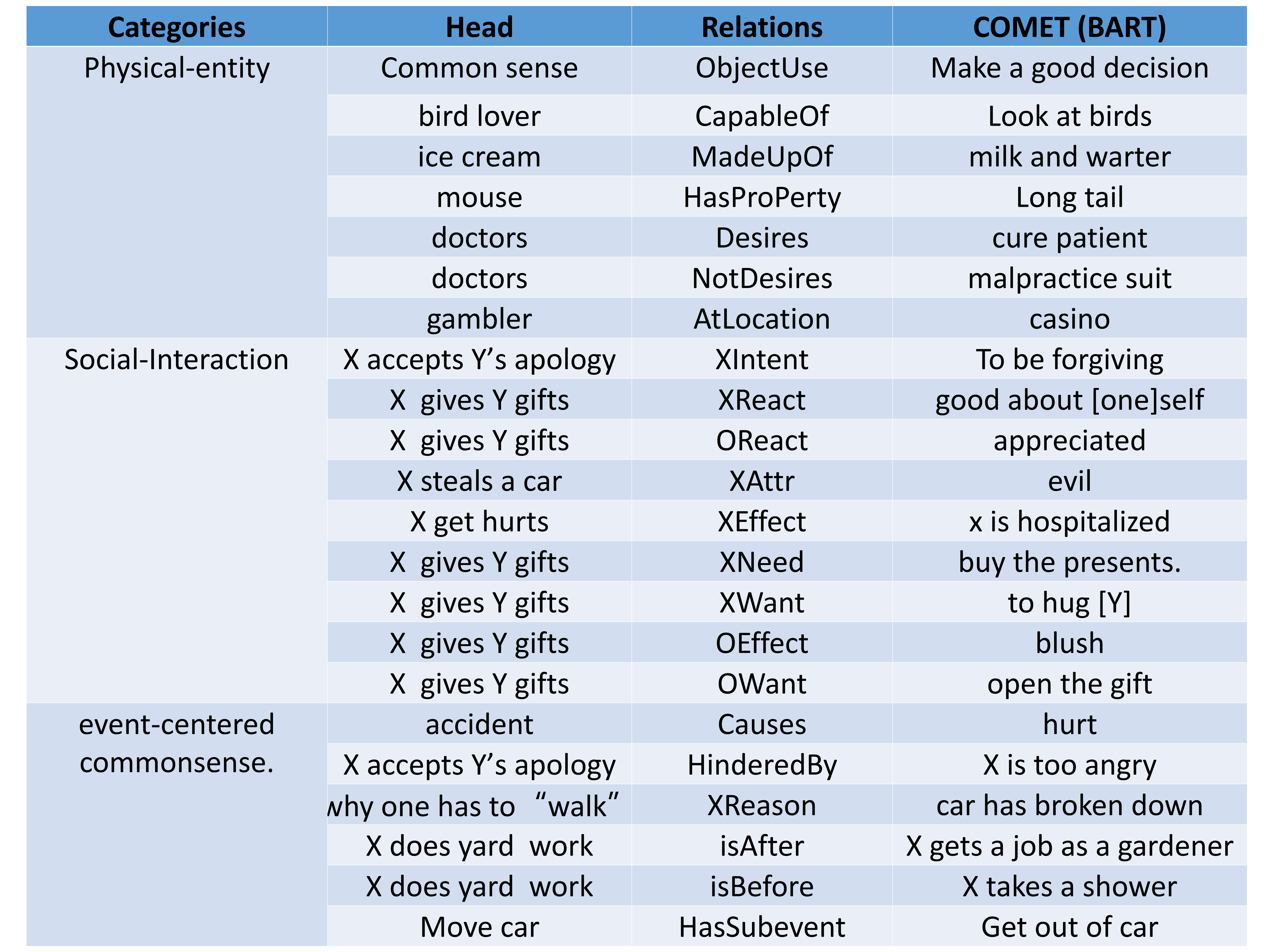}} 
 \caption{Example generations of models on relations from ATOMIC-2020 dataset \citep{hwang2020}.} 
 \label{sec:atomic}
 \end{figure*}

\begin{figure*}[htbp] 
 \center{\includegraphics[width=16cm]  {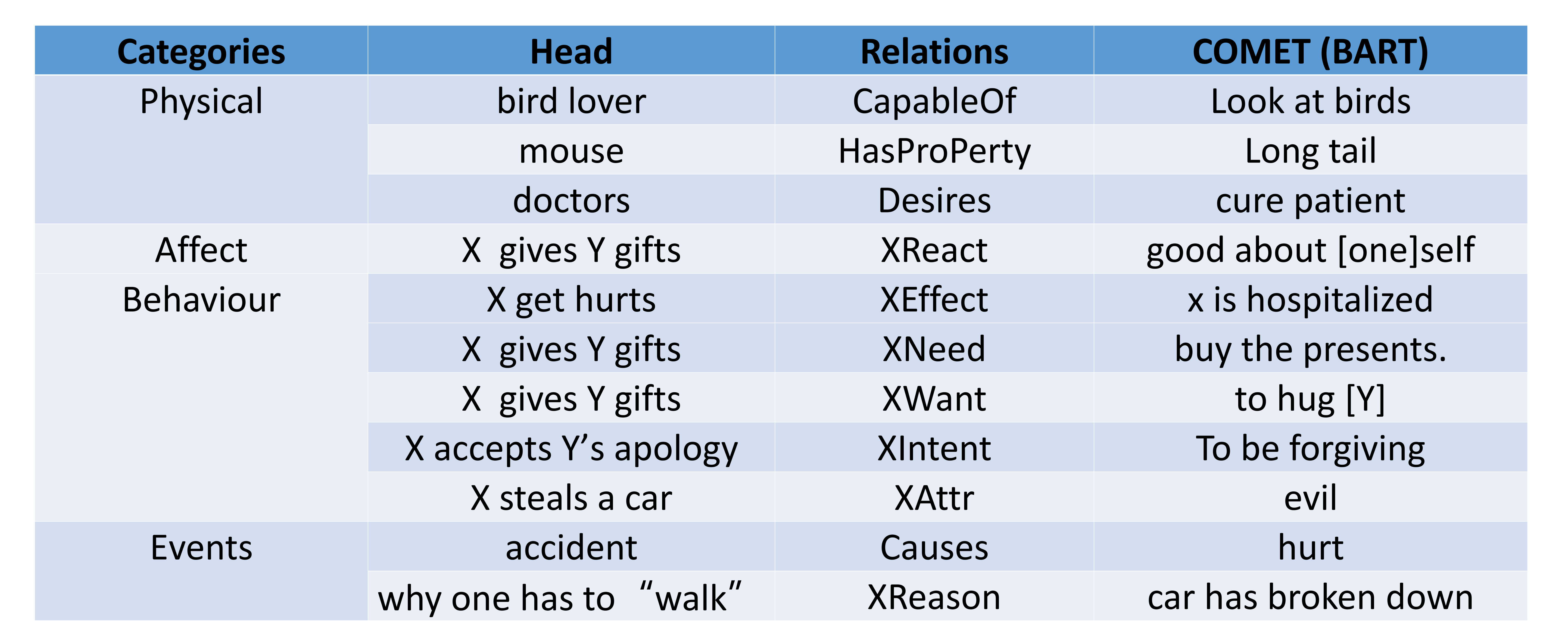}} 
 \caption{Our knowledge graph, which uses 11 relationships and is inspired by psychology, is divided into four modules: Physical, Affect, Behaviour, Events} 
 \label{sec:oursgraph}
 \end{figure*}

In this work, we use the ATOMIC-2020 dataset \citep{hwang2020} as our commonsense knowledge base, which is a collection of commonsense reasoning inferences about everyday if-then contexts. 
They fall into three natural categories based on their meaning: physical-entity, social-interaction, and event-centered commonsense, which are 22 relationships under three categories ( e.g., XReact, XWant, XReason, CapableOf) (See Fig ${\color{red}{\fbox{\ref{sec:atomic}}}}$).
Based on the given contexts, we select those that are speaker-centered and contribute positively to the speaker.
We neglect (oReact,oEffect,oWant, Etc.) in our work. 
Finally, We have extracted 11 important relationships from ATOMIC. 
These relationships are divided into four modules, which are Physical (CapableOf, HasProperty, Desires), Affect (XReact), Behaviour (XEffect, XNeed, XWant, XIntent, XAttr), Events (Causes, XReason).
As shown in Fig ${\color{red}{\fbox{\ref{sec:oursgraph}}}}$.

\section{Implementation Details}
\label{sec:expriments}

Our models are implemented using Pytorch, a modularized, versatile, and extensible toolkit for machine learning and
text generation tasks. 
We used 300-dimensional word embedding and 300-dimensional hidden size everywhere
in our experiments. 
The word embedding is initialized using pre-trained Glove vectors. 
We initialize the transformer encoder with one layer and two attention heads for the task. 
We train our models using Adam optimization with a learning rate of 0.0001. 
Early stopping is applied during training. 
We use a batch size of 1 and a maximum of 30 decoding steps during testing and inference.

\end{document}